\pdfoutput=1

\documentclass[11pt]{article}

\usepackage[final]{EACL2023}

\usepackage{times}
\usepackage{latexsym}

\usepackage{booktabs}

\usepackage[T1]{fontenc}

\usepackage[utf8]{inputenc}

\usepackage{microtype}

\usepackage{inconsolata}

\usepackage{xcolor}
\usepackage{latexsym}

\usepackage{microtype}
\usepackage{amsfonts}
\usepackage{amsmath}
\usepackage{bbm}
\usepackage{graphicx}
\usepackage{multicol}
\usepackage{multirow}
\usepackage{float}

\title{ViDeBERTa: A powerful pre-trained language model for Vietnamese}

\author{
Cong Dao Tran \thanks{$^*$: Co-first authors. $\dagger$: Correspondent author.}\\
FPT AI Center \\
\tt{\normalsize{daotc2@fsoft.com.vn}} \\
\And
Nhut Huy Pham \footnotemark[1]\\
FPT AI Center \\
\tt{\normalsize{huypn10@fsoft.com.vn}} \\
\And
Anh Nguyen \\
Microsoft \\
\tt{\normalsize{anhnguyen@microsoft.com}}
\AND
Truong Son Hy \footnotemark[2]\\
University of California San Diego \\
\texttt{tshy@ucsd.edu}
\And 
Tu Vu \\
University of Massachusetts Amherst \\
\texttt{tuvu@cs.umass.edu}
}

\begin{document}
\maketitle
\begin{abstract}
This paper presents ViDeBERTa, a new pre-trained monolingual language model for Vietnamese, with three versions - $\text{ViDeBERTa}_{xsmall}$, $\text{ViDeBERTa}_{base}$, and $\text{ViDeBERTa}_{large}$, which are pre-trained on a large-scale corpus of high-quality and diverse Vietnamese texts using DeBERTa architecture. Although many successful pre-trained language models based on Transformer have been widely proposed for the English language, there are still few pre-trained models for Vietnamese, a low-resource language, that perform good results on downstream tasks, especially Question answering. We fine-tune and evaluate our model on three important natural language downstream tasks, Part-of-speech tagging, Named-entity recognition, and Question answering. The empirical results demonstrate that ViDeBERTa with far fewer parameters surpasses the previous state-of-the-art models on multiple Vietnamese-specific natural language understanding tasks. Notably, $\text{ViDeBERTa}_{base}$ with 86M parameters, which is only about 23\% of $\text{PhoBERT}_{large}$ with 370M parameters, still performs the same or better results than the previous state-of-the-art model. Our ViDeBERTa models are available at: \url{https://github.com/HySonLab/ViDeBERTa}.

\end{abstract}

\section{Introduction}

In recent years, pre-trained language models (PLMs) and Transformer-based architecture models have been essential in the advancement of Natural Language Processing (NLP). Large-scale Transformer-based pre-trained models with the capacity to derive a contextual representation of the languages in the training data include GPT \cite{radford2019language, brown2020language}, BERT \cite{devlin2019bert}, RoBERTa \cite{liu2019roberta}, XLNet \cite{yang2019xlnet}, ELECTRA \cite{clark2020electra}, T5 \cite{raffel2020exploring}, and DeBERTa \cite{he2020deberta, he2021debertav3}. Following pre-training, these models performed at the cutting edge on various downstream NLP tasks \cite{devlin2019bert}. The development of pre-trained models in other languages, including Vietnamese (PhoBERT \cite{nguyen2020phobert}; ViBERT \cite{tran2020improving}; ViT5 \cite{phan2022vit5}), and Arabic \cite{antoun2021aragpt2}, has been spurred on by the success of pre-trained models in English. In order to enhance performance across several languages by learning both general and language-specific representations, multilingual pre-trained models  ( XLM-R \cite{conneau2020unsupervised}, mT5 \cite{xue2021mt5}, and mBART \cite{liu2020multilingual} are also being developed.

Most recently, PhoBERT \cite{nguyen2020phobert}, the first large pre-trained model for Vietnamese that inherits the RoBERTa \cite{liu2019roberta} architecture, has demonstrated the effectiveness of the trained language model compared with current methods modernized in four Vietnamese-specific tasks, including Part of Speech Tagging (POS), Dependency Parsing, Named Entity Recognition (NER), and Natural Language Inference (NLI). 
Nevertheless, there are still rooms to build an improved pre-trained language model for Vietnamese. Firstly, PhoBERT was pre-trained on a relatively small Vietnamese dataset of 20GB of uncompressed texts, while pre-trained language models can be significantly improved by using more pre-training data \cite{liu2019roberta}. Secondly, Question answering (QA) is one of the most impactful tasks that has mainly focused on the computational linguistics and artificial intelligence research community within information retrieval and information extraction in recent years. However, there are a few pre-trained models for Vietnamese that produce efficient results in the QA tasks, especially PhoBERT \cite{nguyen2020phobert} and ViT5 \cite{phan2022vit5}. Last but not least, some previous works point to DeBERTa architecture \cite{he2020deberta, he2021debertav3} using several novel techniques that can significantly outperform RoBERTa and improve the efficiency of model pre-training and the performance of downstream tasks in some respects. 

Inspired by that, we introduce an improved large-scale pre-trained language model, ViDeBERTa, trained on CC100 Vietnamese monolingual, following the architecture and pre-training methods of DeBERTaV3 \cite{he2021debertav3}. We comprehensively evaluate and compare our model with competitive baselines, i.e., the previous SOTA models PhoBERT, ViT5, and the multilingual model XLM-R  on three Vietnamese downstream tasks, including POS tagging, NER, and QA. In this work, we focus on two main categories of QA: Machine Reading Comprehension (MRC) and Open-domain Question Answering (ODQA). The experiment results show the performance of our model surpasses all baselines on all tasks. 
Our main contributions are summarized as follows:
\begin{itemize}
    \item We present and implement ViDeBERTa with three versions: $\text{ViDeBERTa}_{xsmall}$, $\text{ViDeBERTa}_{base}$, and $\text{ViDeBERTa}_{large}$ which are the improved large-scale monolingual language models pre-trained for Vietnamese based on the DeBERTa architecture and pre-training procedure.
    \item We also conduct extensive experiments to verify the performance of our pre-trained models compared to previous strong models in terms of Vietnamese language modeling.
    Our empirical results demonstrated the state-of-the-art (SOTA) results on Vietnamese downstream tasks: POS tagging, NER, and QA, thus confirming the effectiveness of our improved pre-trained language model for Vietnamese.
    \item Our model, ViDeBERTa, which works with \textit{huggingface} and \textit{transformers}, is available to the public. We expect that ViDeBERTa will be an effective pre-trained model for many NLP applications and research in Vietnamese and other low-resource languages.
\end{itemize}
\section{Related work}
\textbf{Pre-trained language models for Vietnamese.} 
PhoBERT \cite{nguyen2020phobert} is the first large-scale PLM for Vietnamese, which has the same architecture as BERT \cite{devlin2019bert} and the same pre-training approach as RoBERTa \cite{liu2019roberta} for more robust performance. This model was trained on a Vietnamese Wikipedia corpus of 20GB word-level texts and produced SOTA results on Vietnamese understanding tasks such as POS, NER, Dependency parsing, and NLI. Following PhoBERT, ViBERT \cite{tran2020improving} and ViELECTRA are public monolingual language models for Vietnamese based on BERT and ELECTRA pre-training techniques \cite{clark2020electra} that are pre-trained on syllable-level Vietnamese textual data.
Recent works such as BARTpho \cite{tran2021bartpho} and ViT5 \cite{tran2020improving} are pre-trained for Vietnamese text summarization. 

\noindent
\textbf{Fine-tuning tasks.}
This work utilizes three Vietnamese natural language understanding (NLU) tasks, including POS tagging, NER, and QA, for fine-tuning and evaluating our model's performance. For POS tagging and NER, PhoBERT still produces better results than ViELECTRA, PhoNLP, and ViT5 \cite{nguyen2020phobert, nguyen2021phonlp, phan2022vit5}. While early QA \cite{voorhees1999trec, brill2002analysis, ferrucci2010building} systems were commonly complex and had many parts, MRC models have evolved and now suggest a simpler two-stage retriever-reader framework \cite{chen2017reading}. A context retriever first selects a small subset of passages where some of them contain the answer to the question then a machine reader can carefully review the retrieved contexts and determine the correct answer. The tasks based on QA have gained much attention in recent years in the Vietnamese natural language processing and computational linguistics community. However, to the best of our knowledge, there is only the work \cite{van2022xlmrqa} that proposes the first Vietnamese retriever-reader QA system employing a transformer-based model (XLM-R) evaluated on the ViQuAD corpus \cite{nguyen2020vietnamese}.

\section{ViDeBERTa}
\label{sec: model}

\subsection{Pre-training data}

In this work, we use a large corpus CC100 Dataset of 138GB uncompressed texts (Monolingual Datasets from Web Crawl Data) \cite{conneau2020unsupervised} as a pre-training dataset.
This corpus includes data for romanized languages and monolingual data for more than 100 languages.

According to \citet{nguyen2020phobert, tran2021bartpho}, pre-trained language models trained on word-level data can perform better than those trained on syllable-level data for word-level Vietnamese NLP tasks. As a result, we perform word and sentence segmentation using a Vietnamese toolkit PyVi \footnote{https://pypi.org/project/pyvi/} on the pre-training dataset. After that,  we use a pre-trained SentencePiece tokenizer from DeBERTaV3 \cite{he2021debertav3} to segment these sentences with sub-word units, which have a vocabulary of 128K sub-word types.


\subsection{Model Architecture}

Our model, ViDeBERTa, follows the DeBERTaV3 architecture by \citet{he2021debertav3}, which is trained using the self-supervise learning objectives of MLM and RTD task and a new weight-sharing Gradient-Disentangled Embedding Sharing (GDES) to enhance the performance of the model. We present three versions of our model, $\text{ViDeBERTa}_{xsmall}$, $\text{ViDeBERTa}_{base}$, and $\text{ViDeBERTa}_{large}$  with 22M, 86M, and 304M backbone parameters, respectively. 

The details of our model architecture hyper-parameters are listed in Table \ref{tab:model_hyperparam}.

\begin{table}[htbp]
  \centering
  \caption{Statistic of our model hyper-parameters. \#layer and \#heads denote the numbers of layers and attention heads of ViDeBERTa model versions, respectively.}
  \scalebox{0.92}{
    \begin{tabular}{lccc}
    \toprule
    Model & \#layers & \#heads & hidden size \\
    \midrule
    $\text{ViDeBERTa}_{xsmall}$ & 6     & 12    & 768 \\
    $\text{ViDeBERTa}_{base}$ & 12    & 12    & 768 \\
    $\text{ViDeBERTa}_{large}$ & 24    & 12     & 1024 \\
    \bottomrule
    \end{tabular}%
  }

  \label{tab:model_hyperparam}%
\end{table}%

\subsection{Optimization}

We employ our model based on the DeBERTaV3 implementation from \cite{he2021debertav3}. We use Adam \cite{kingma2015adam} as the optimizer with weight decay \cite{loshchilov2018decoupled}
and use a global batch size of 8,192 across 32 A100 GPUs (80GB each) and a peak learning rate of 6e-4 for both $\text{ViDeBERTa}_{xsmall}$ and $\text{ViDeBERTa}_{base}$, while peak learning rate of 3e-4 was used for $\text{ViDeBERTa}_{large}$. We pre-train $\text{ViDeBERTa}_{xsmall}$ and $\text{ViDeBERTa}_{base}$ for 500k training iterations and $\text{ViDeBERTa}_{large}$ for 250k training iterations.

\section{Experiments and Results}
\label{sec: experiment_result}

\subsection{POS tagging and NER}
\subsubsection{Experimental setup}

For POS tagging and NER tasks, we use standard benchmarks of the VLSP POS tagging dataset \footnote{https://vlsp.org.vn/vlsp2013/eval/ws-pos} and the PhoNER dataset \cite{truong2021covid}.
We follow the procedure in \citealt{devlin2019bert, nguyen2020phobert} to fine-tune our pre-trained model for POS tagging and NER tasks. In particular, a linear layer for prediction is appended on top of our model architecture (the last Transformer layer). We then use Adam \cite{kingma2015adam} to optimize our model for fine-tuning with a fixed learning rate of 1e-5 and batch size of  16 \cite{he2021debertav3}. 
The final results for each task and each dataset are averaged and reported over five independent runs with different random seeds.

We compare the performance of ViDeBERTa models with the solid baselines, including PhoBERT, XLM-R, and ViT5, for these tasks. Here, XLM-R is a multilingual masked language model pre-trained on 2.5 TB of CommmonCrawl dataset of 100 languages, which includes 137GB of Vietnamese texts.

\subsubsection{Main results}

\begin{table}[htbp]
  \centering
   
  \scalebox{1}{
    \begin{tabular}{lccc}
    \hline
    \multicolumn{1}{l}{\multirow{2}[4]{*}{\textbf{Model}}} & \textbf{POS} & \textbf{NER} &  \textbf{MRC}\\
\cline{2-4}          & Acc. & $\text{F}_1$ & $\text{F}_1$\\
    \hline
    $\text{XLM-R}_{base}$ & $96.2^\dagger$  & \_ & $82.0^\ddagger$\\
    $\text{XLM-R}_{large}$ & $96.3^\dagger$  & $93.8^\star$ & $87.0^\ddagger$\\
    $\text{PhoBERT}_{base}$ & $96.7^\dagger$  & $94.2^\star$ & 80.1\\
    $\text{PhoBERT}_{large}$ & $96.8^\dagger$  & $94.5^\star$ & 83.5\\
    $\text{ViT5}_{base 1024-length} $&  \_     & $94.5^\star$ &  \_ \\ 
    $\text{ViT5}_{large 1024-length}$ &  \_     & $93.8^\star$ & \_  \\
    \hline
    $\text{ViDeBERTa}_{xsmall}$ &  96.4     & 93.6 & 81.3\\
    $\text{ViDeBERTa}_{base}$ &   96.8    & 94.5 & 85.7 \\
    $\text{ViDeBERTa}_{large}$ &  \textbf{97.2}    & \textbf{95.3}& \textbf{89.9}\\
    \hline
    \end{tabular}%
  }
  \caption{Test results (\%) for three tasks POS tagging (POS for short), NER, and MRC on test sets. Note that ``Acc.'' abbreviates the accuracy. $\dagger$, $\star$, and $\ddagger$ denote scores taken from the PhoBERT paper \cite{nguyen2020phobert}, the ViT5 paper \cite{phan2022vit5}, and the ViQuAD paper \cite{nguyen2020vietnamese}, respectively.}
  \label{tab:pos_ner_mrc}
\end{table}%
Table \ref{tab:pos_ner_mrc} shows the obtained scores of ViDeBERTa compared to the baselines with the highest reported results. It can be seen clearly that our model produces significantly better results than the baselines and achieves new SOTA performance on both POS tagging and NER tasks.

For POS tagging, ViDeBERTa obtains 0.9\% and 0.4\% absolute higher accuracy than the large-scale multilingual model XLM-R \cite{nguyen2020vietnamese} and the previous SOTA model PhoBERT \cite{nguyen2020phobert}, respectively
. Table \ref{tab:pos_ner_mrc} also shows our $\text{ViDeBERTa}_{xsmall}$ obtains 96.4\% accuracy that are better than the baseline $\text{XLM-R}_{large}$ and $\text{ViDeBERTa}_{base}$ obtains 96.8\% that are competitively the same as the $\text{PhoBERT}_{large}$.

For NER, our $\text{ViDeBERTa}_{large}$ achieves $\text{F}_1$ score at 95.3\% and improves 0.8\% absolute higher score than the previous SOTA models $\text{ViT5}_{base 1024-length}$ and $\text{PhoBERT}_{large}$. Furthermore, $\text{ViDeBERTa}_{large}$ and $\text{ViDeBERTa}_{base}$ preform 1.5\% and 0.7\% absolute higher scores than the baseline $\text{XLM-R}_{large}$ on the PhoNER corpus.

\subsection{Question Answering}
\subsubsection{Experimental setup}

\begin{figure*}[!ht]
    \centering
    \includegraphics[scale = 0.66]{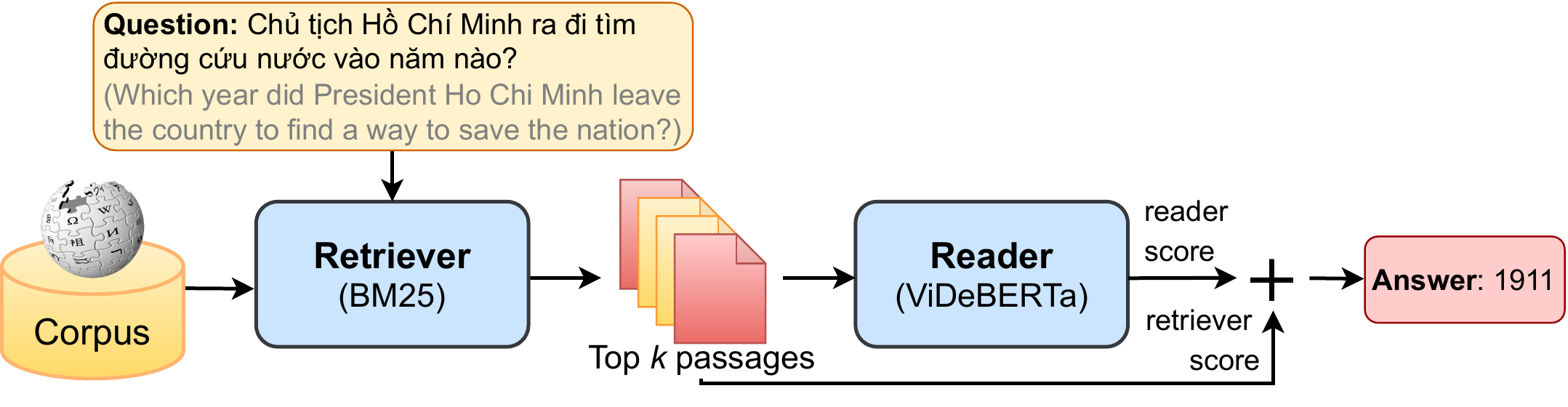}
    \caption{An overview of ViDeBERTa-QA framework for Vietnamese Open-domain Question Answering task.}
    \label{fig:ViDeBERTa-QA_architecture}
\end{figure*}
For QA, we evaluate our model on two main tasks: MRC and ODQA.
For ODQA, we propose a new framework ViDeBERTa-QA, that uses a BM25 \cite{robertson2009probabilistic}  as a retriever and ViDeBERTa as a text reader. 

Figure \ref{fig:ViDeBERTa-QA_architecture} depicts an overview of our ViDeBERTa framework for the Vietnamese Open-domain Question answering task.
The statistics of the ViQuAD dataset used for the task, which is introduced by \citet{nguyen2020vietnamese}, are summarized in Table \ref{tab:viquad}.

\begin{table}[htbp]
  \centering
  
    \begin{tabular}{lccc}
    \hline
    Corpus & \#article & \#passage & \#question\\
    \hline
    Train & 138   & 4,101 & 18,579 \\ 
    Dev   & 18    & 515   & 2,285 \\
    Test  & 18    & 493   & 2,21 \\
    \hline
    Full  & 174   & 5,109 & 23,074 \\
    \hline
    \end{tabular}%
    \caption{Statistics of the ViQuAD dataset for QA. ``\#article'', ``\#valid'', and ``\#test'' denote the number of articles, passages, and questions in the ViQuAD, respectively.}
  \label{tab:viquad}
\end{table}%

We compare ViDeBERTa to the best model XLM-R \cite{nguyen2020vietnamese} and PhoBERT \footnote{We carefully fine-tune PhoBERT for the MRC task following the fine-tuning approach that we use for ViDeBERTa.} for Vietnamese MRC. We also compare our framework, ViDeBERTa-QA, to strong baselines DrQA \cite{chen2017reading}, BERTserini \cite{yang2019end}, and the first Vietnamese ODQA system XLMRQA \cite{van2022xlmrqa}) that uses $\text{XLM-R}_{large}$ as a reader. 
We use the ViQuAD corpus introduced by \citet{nguyen2020vietnamese} for assessing these tasks. ViQuAD is a Vietnamese corpus that comprises over 23k triples and each triple includes a question, its answer, and a passage containing the answer.

Similar to POS tagging and NER, we use Adam \cite{kingma2015adam} as an optimizer with a learning rate of 2e-5 and a batch size of 16. We report the final results as an average over five independent runs with different random seeds.

\subsubsection{Main results}
Table \ref{tab:pos_ner_mrc} presents the results obtained by ViDeBERTa and two baselines XLM-R (reported by \citet{nguyen2020vietnamese}) and PhoBERT for MRC on ViQuAD corpus. We find that our ViDeBERTa performance outperforms both XLM-R and PhoBERT in terms of $\text{F}_1$ score.

In particular, the previous SOTA model $\text{XLM-R}_{large}$ for Vietnamese MRC obtains 87\%. Clearly, ViDeBERTa helps boost the XLM-R with about 2.9\% absolute improvement, obtaining a new SOTA result at 89.9\%. In addition, both versions $\text{ViDeBERTa}_{base}$ and $\text{ViDeBERTa}_{large}$ also outperform $\text{PhoBERT}_{base}$ and $\text{PhoBERT}_{large}$ by large margins, respectively. Especially, $\text{ViDeBERTa}_{xsmall}$ (22M parameters) produces 1.2\% absolute higher score than $\text{PhoBERT}_{base}$ (135M parameters) and $\text{ViDeBERTa}_{base}$ (86M parameters) produces 2.2\% absolute higher score than $\text{PhoBERT}_{large}$ (370M parameters) but uses far fewer parameters than PhoBERT.

For ODQA, Table \ref{tab:opqa} shows the obtained $\text{F}_1$ scores for ViDeBERTa-QA and its baselines on the test set. Obviously, ViDeBERTa-QA achieves better scores than the previous SOTA XLMRQA, BERTsini, and DrQA at the top $k$ passages, selected by retrievers, is 10 and 20. In particular, ViDeBERTa-QA performs 0.85\% (at $k=20$) and 0.4\% (at $k=10$) absolute higher scores than the previous SOTA system. At smaller $k$ (= 1, 5), ViDeBERTa performs better BERTserini and DrQA by a large margin; however, XLMRQA does better than ViDeBERTa-QA.
\begin{table}[htbp]
  \centering

    \scalebox{0.88}{
     \begin{tabular}{lcccc}
        \hline
        \multicolumn{1}{l}{\multirow{2}[4]{*}{\textbf{Model}}} & \multicolumn{4}{c}{Top $k$ selected passages} \\
        \cline{2-5}          & \textbf{1} & \textbf{5} & \textbf{10} & \textbf{20} \\
        \hline

        DrQA [*] & 37.86  & 37.86   & 37.86   & 37.86  \\
        BERTserini [*] & 55.55 & 58.30 & 57.98    & 58.09 \\
        XLMRQA [*] & \textbf{61.83} & \textbf{64.99}    & 64.49 & 64.49 \\
        \hline
        $\text{ViDeBERTa}_{xsmall}$ &  52.76     &  56.24     &  56. 93    &  57.40 \\
        $\text{ViDeBERTa}_{base}$ &   58.55    &  61.37     &  61.89     & 62.43 \\
        $\text{ViDeBERTa}_{large}$ &   61.23    &  63.57     &   \textbf{64.89}    &  \textbf{65.34} \\
        \hline
    \end{tabular}%
    }
     \caption{Test scores ($\text{F}_1$ in \%) for ODQA on ViQuAD corpus with different $k$ values. Note that [*] indicates the results reported following \citet{van2022xlmrqa}.}
  \label{tab:opqa}
\end{table}%

\subsection{Discussion}
According to the results on both downstream tasks of POS tagging and NER in Table \ref{tab:pos_ner_mrc}, we find that $\text{ViDeBERTa}_{xsmall}$ (86M) with fewer parameters (i.e. only about 15\% of $\text{XLM-R}_{large}$ 560M and 25\% of $\text{PhoBERT}_{large}$ 370M) but still performs slightly better than $\text{XLM-R}_{large}$ and competitively the same as the previous SOTA $\text{PhoBERT}_{large}$. One possible reason is that our model inherits the robustness of DeBERTaV3 architecture and pre-training techniques, which are demonstrated superior performance by \citet{he2020deberta, he2021debertav3}. Moreover, using more high-quality pre-training data (138GB) can help ViDeBERTa  significantly improve its performance compared to PhoBERT (using 20GB). 

For Vietnamese QA, the results on the MRC task show that ViDeBERTa outperforms PhoBERT by a large margin. It is worth noting that PhoBERT set a maximum length of 256 subword tokens for both versions while ViDeBERTa set a larger one of 512. As a result, our models are more scalable than PhoBERT for long contexts. The results obtained by ViDeBERTa-QA on ODQA also suggest that our framework achieves the best performance with large top $k$ passages selected by the retriever (i.e. $k = 10, 20$).

\label{sec: discussion}

\section{Conclusion}
\label{sec: conclusion}
In this paper, we have introduced ViDeBERTa, a new pre-trained large-scale monolingual language model for Vietnamese. We demonstrate the effectiveness of our ViDeBERTa by showing that ViDeBERTa with fewer parameters performs better than the recent strong pre-trained language models as XLM-R, PhoBERT, and ViT5, and achieves SOTA performances for three downstream Vietnamese language understanding tasks, including POS tagging, NER, and especially QA.  We hope that our public ViDeBERTa model will boost ongoing NLP research and applications for Vietnamese and other low-resource languages.

\section*{Limitations}
\label{sec:limitations}


While we have shown that ViDeBERTa can achieve state-of-the-art performance on a variety of NLP tasks for Vietnamese, we believe that more analyses and ablations are required to better understand what facets of ViDeBERTa contributed to its success and what knowledge of Vietnamese that ViDeBERTa captures. We leave these further explorations to future work.


\bibliographystyle{acl_natbib}
\bibliography{paper}

\clearpage
\appendix

\section{Background of DeBERTa}


DeBERTa enhances BERT with disentangled attention and a more powerful mask decoder. The disentangled attention mechanism is distinct from prior methods in that it uses two distinct vectors to represent each input word: one for the content and one for the location. The words' attention weights are calculated using disentangled matrices based on both their relative placements and contents. Similar to BERT, DeBERTa has been pre-trained using masked language modeling. The disentangled attention process already accounts for the relative locations and contents of the context words but not for their absolute positions, which are usually crucial for prediction. DeBERTa improves MLM by utilizing a better mask decoder at the MLM decoding layer and absolute position information of the context words.

\subsection{Masked Language model}

Large-scale Transformer-based PLMs are often pre-trained using a self-supervision aim called Masked Language Model (MLM) \cite{devlin2019bert} to learn contextual word representations in enormous volumes of text. In further detail, we corrupt a given sequence $\textbf{\textit{X}} = \{x_i\}$ into $\tilde{\textbf{\textit{X}}}$ by randomly masking 15\% of its tokens and train a language model parameterized by $\theta$  to reconstruct $\textbf{\textit{X}}$ by anticipating the masked tokens $\tilde{x}$ conditioned on $\tilde{\textbf{\textit{X}}}$:

\begin{equation}
    \max_\theta \log p_\theta(\textbf{\textit{X}}|\tilde{\textbf{\textit{X}}}) = \max_\theta \sum_{i\in C}\log p_\theta(\tilde{x_i} = x_i| \tilde{\textbf{\textit{X}}}),
\end{equation}
where $C$ is the sequence's index set for the masked tokens. The authors of BERT suggest keeping 10\% of the masked tokens unchanged, replacing another 10\% with tokens chosen at random, and replacing the remaining tokens with the [MASK] token.

\subsection{Replaced token detection}
Like ELECTRA, which was trained with two transformer encoders in GAN style, DeBERTaV3 \cite{he2021debertav3} improves DeBERTa by using the training loss in the generator is MLM and discriminator is Replaced Token Detection (RTD). The loss function of the generator can be written as follows:
\begin{equation}
    L_{MLM} =  \mathbb{E}\left(-\sum_{i\in C}\log p_{\theta_G}(\tilde{x}_{i,G} = x_i|\tilde{\textbf{\textit{X}}}_G) \right),
\end{equation}
where $\theta_G$ and $\tilde{\textbf{\textit{X}}}_G$ are the parameter and the input of the generator by masking $15\%$ tokens in $\textbf{\textit{X}}$, respectively.

The discriminator's input sequence is constructed by replacing masked tokens with new tokens sampled according to the generator's output probability:

\begin{equation}
    \tilde{x}_{i,D} = 
		\begin{cases} \tilde{x}_i \sim p_{\theta_G}(\tilde{x}_{i,G} = x_i|\tilde{\textbf{\textit{X}}}_G), & i \in C
		\\ 
        x_i, & i \notin C
		\end{cases}
\end{equation}
The loss function of the discriminator is written as follows:

\begin{equation}
    \label{eq:L_rtd}
     L_{RTD} =  \mathbb{E}\left(-\sum_{i}\log p_{\theta_G}(\mathbbm{1}(\tilde{x}_{i,D} = x_i)|\tilde{\textbf{\textit{X}}}_D) \right),
\end{equation}
where $\theta_D$ is the parameter of the discriminator, $\mathbbm{1}(\cdot)$ is the indicator function, and $\tilde{\textbf{\textit{X}}}_D$ is the input to the discriminator constructed by Equation \ref{eq:L_rtd}. Then $L_{MLM}$ and $L_{RTD}$ are optimized jointly by the final loss $L = L_{MLM} +  \lambda L_{RTD}$, where $\lambda$ is the weight of the discriminator loss.
\label{sec:related_work}

Besides using the RTD training loss like ELECTRA \cite{clark2020electra}, DeBERTaV3 improves DeBERTa by using a new weight-sharing method called Gradient-Disentanggled Embedding Sharing (GDES) \cite{he2021debertav3}. The experimental results conducted by \citeauthor{he2021debertav3} indicate that GDES is an effective weight-sharing method for language model pre-trained with MLM and RTD tasks.

\end{document}